\documentclass[letterpaper]{article} 
\usepackage{aaai25}  
\usepackage{times}  
\usepackage{helvet}  
\usepackage{courier}  
\usepackage[hyphens]{url}  
\usepackage{graphicx} 
\usepackage{natbib}  
\usepackage{caption} 
\usepackage{algorithm}
\usepackage{algorithmic}

\usepackage{multirow, array, booktabs,makecell,xspace,bm}
\usepackage{arydshln}
\usepackage{amsmath}
\usepackage{amsfonts}
\usepackage{amssymb}

\usepackage{mathtools}
\usepackage{pifont}
\usepackage{dsfont}

\newcommand{\cmark}{\ding{51}}%
\usepackage{newfloat}
\usepackage{listings}
\DeclareCaptionStyle{ruled}{labelfont=normalfont,labelsep=colon,strut=off} 
\floatstyle{ruled}
\newfloat{listing}{tb}{lst}{}
\floatname{listing}{Listing}
\title{Watch Video, Catch Keyword: Context-aware Keyword Attention\\for Moment Retrieval and Highlight Detection}

\author {
    Sung Jin Um\textsuperscript{\rm 1},
    Dongjin Kim\textsuperscript{\rm 1},
    Sangmin Lee\textsuperscript{\rm 2,}\equalcontrib,
    Jung Uk Kim\textsuperscript{\rm 1,}\equalcontrib
}
\affiliations {
    \textsuperscript{\rm 1}Kyung Hee University, Yong-in, South Korea\\
    \textsuperscript{\rm 2}Sungkyunkwan University, Seoul, South Korea\\
    \{sungzin1, rlaehdwls310, ju.kim\}@khu.ac.kr, sangmin.lee@skku.edu
}

\def\maketitlesupplementary
   {
    \newpage
       \twocolumn[
        \centering
        \Large
        \textbf{Watch Video, Catch Keyword: Context-aware Keyword Attention \\ for Moment Retrieval and Highlight Detection\\--\textit{Supplementary Material}--}\\
        \vspace{1.5em}
       ] 
   }

\begin{document}

\maketitle

\begin{abstract}
The goal of video moment retrieval and highlight detection is to identify specific segments and highlights based on a given text query. With the rapid growth of video content and the overlap between these tasks, recent works have addressed both simultaneously. However, they still struggle to fully capture the overall video context, making it challenging to determine which words are most relevant. In this paper, we present a novel Video Context-aware Keyword Attention module that overcomes this limitation by capturing keyword variation within the context of the entire video. To achieve this, we introduce a video context clustering module that provides concise representations of the overall video context, thereby enhancing the understanding of keyword dynamics. Furthermore, we propose a keyword weight detection module with keyword-aware contrastive learning that incorporates keyword information to enhance fine-grained alignment between visual and textual features. Extensive experiments on the QVHighlights, TVSum, and Charades-STA benchmarks demonstrate that our proposed method significantly improves performance in moment retrieval and highlight detection tasks compared to existing approaches. Our code is available at: \url{https://github.com/VisualAIKHU/Keyword-DETR}.
\end{abstract}

%

\section{Introduction}

With the exponential growth of video content, precise video moment retrieval and highlight detection have become crucial \cite{snoek2009concept:intro, apostolidis2021video:intro}. Video moment retrieval enables users to find specific segments within videos based on natural language queries \cite{anne2017localizing:mr_intro}, while highlight detection helps extract the most engaging parts from long-form videos \cite{gygli2014creating:hd_intro}. These technologies enhance user experience and productivity across various applications, including video searching, video editing, social media, and e-learning, by enabling quick and accurate access to relevant content.

Extensive research has been conducted on moment retrieval \cite{gao2017tall:charades-sta, hendricks2018localizing:mr, xiao2021boundary:mr, sun2022you:mr} and highlight detection \cite{sun2014ranking:hd, xu2021cross:hd, wei2022learning:hd, badamdorj2022contrastive:hd} as separate tasks. However, with the introduction of Moment-DETR and the QVHighlights dataset \cite{{moment_detr2021:mr_hd}}, which allows for the simultaneous execution of these tasks, new studies \cite{umt2022:mr_hd, qd_detr2023:mr_hd, uvcom2024:mr_hd, tr_detr2024:mr_hd} have emerged that aim to address moment retrieval and highlight detection based on text queries concurrently. Following the Moment-DETR \cite{moment_detr2021:mr_hd}, UMT \cite{umt2022:mr_hd} utilizes audio-visual multi-modal learning. TR-DETR \cite{tr_detr2024:mr_hd} leverages the reciprocal relationship between two tasks, refining visual features through textual guidance to enhance both tasks simultaneously. UVCOM \cite{uvcom2024:mr_hd} introduces integration module for progressive intra- and intermodality interaction across multi-granularity.

\begin{figure}[t]
\begin{minipage}[b]{1.0\linewidth}
\centering
\centerline{\includegraphics[width=0.995\linewidth]{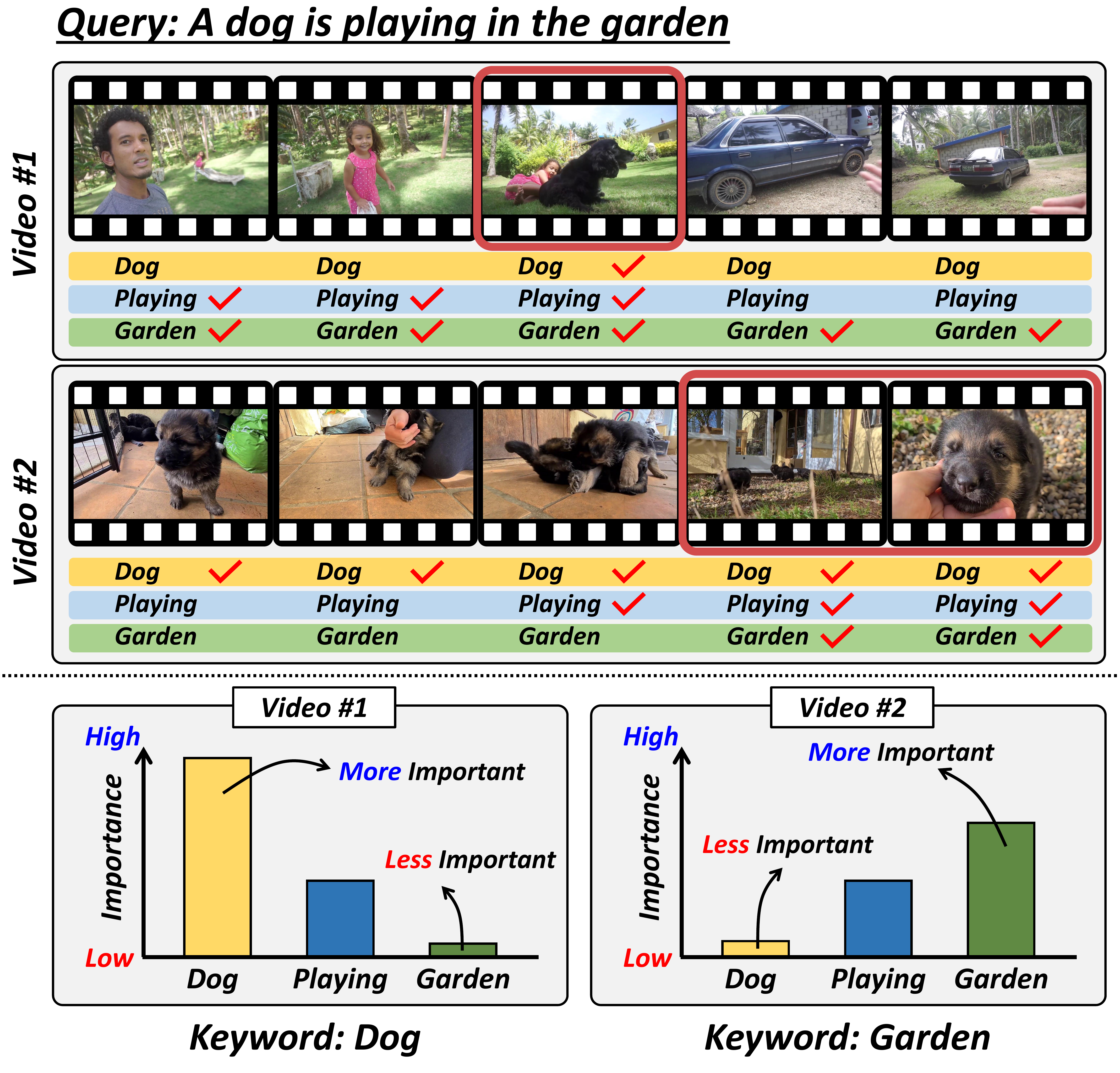}}
\end{minipage}
\caption{Text keywords can vary by video context. The less frequently a word appears in the video clip, the more important it becomes within the text query. In Video \#1, `\textit{dog}' is important, while in Video \#2, `\textit{garden}' is important.}
\label{fig1}
\end{figure}

Despite these advancements, existing methods often fail to capture the dynamic importance of keywords within the context of the video, which is crucial for accurate moment retrieval and highlight detection. As illustrated in Figure \ref{fig1}, the importance of each word in a text query can vary depending on the video content. For instance, when we consider a text query \textit{`A dog is playing in the garden'}, the importance of words such as \textit{`dog'} or \textit{`garden'} can shift based on the predominant scenes in the video. In Video \#1, where most scenes contain \textit{`garden'}, the word \textit{`dog'} is more critical than \textit{`garden'} for specifying the desired video segment. Conversely, in Video \#2, \textit{`dog'} is predominantly \textit{`playing'} indoors, making \textit{`garden'} more essential than \textit{`dog'} for identifying the relevant video segment. This indicates that the importance of words in a text query can vary significantly depending on the video context. Therefore, it is necessary to consider such keyword variations for effective moment retrieval and highlight detection. However, existing methods fall short in addressing this keyword variation, as they rely on text features extracted independently of the video context, failing to capture the dynamic importance of words relative to the visual content.

In this paper, we propose a Video Context-aware Keyword Attention module that effectively captures keyword variations by considering the overall video context. Our approach addresses two main challenges: (\textit{i}) how to effectively encode the overall context of a video to capture keyword variation, and (\textit{ii}) how to capture and utilize desired text keywords within their relevant video contexts.

First, effective keyword extraction requires a comprehensive understanding of the overall context of the video. To address this, we tackle the challenge (\textit{i}) by introducing a video context clustering module that leverages temporally-weighted clustering to group similar video scenes. This approach allows our model to grasp the high-level flow and structure of the video. The resulting cluster assignments provide a concise representation of the overall video context and are leveraged to understand keyword dynamics. Furthermore, since these clustered features contain information about scene changes, they are further used as additional hints for moment retrieval and highlight detection.

To address the challenge (\textit{ii}), we propose a keyword weight detection module. This module recognizes less frequently occurring but important words in the text query and calculates the similarity between clustered video features and text features to generate a keyword weight vector. This vector captures information about the important words in the text query within the video context, allowing our framework to adjust the keywords based on the overall video context dynamically. Based on this, we introduce keyword-aware contrastive learning to incorporate keyword weights and facilitate a fine-grained alignment between visual and text features. As a result, our method allows for accurate moment retrieval and highlight detection.

The major contributions of our paper are as follows:
\begin{itemize}
    \item We propose a video context-aware keyword attention module to capture keyword variations by considering overall context of the video for effective moment retrieval and highlight detection. To the best of our knowledge, this is the first work to address this aspect in video moment retrieval and highlight detection tasks.
    \item We introduce keyword-aware contrastive learning to integrate keyword weight information, enhancing the fine-grained alignment between visual and text features. This approach improves the ability of model to understand the relationship between textual queries and video content.
    \item Experimental results on QVHighlights, TVSum, and Charades-STA demonstrate the effectiveness of our method for moment retrieval and highlight detection.
\end{itemize}

\begin{figure*}[t]
\centering
\includegraphics[width=1.0\textwidth]{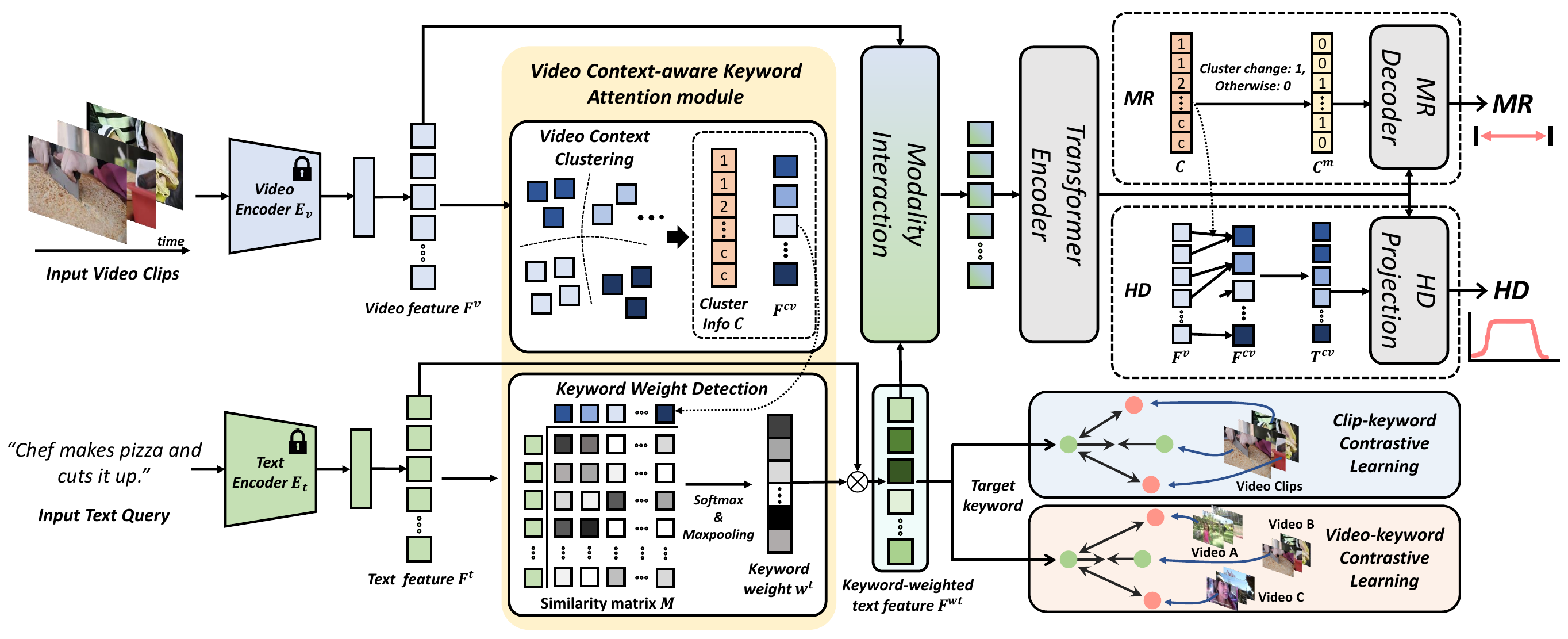}
\caption{Overall configuration of our moment retrieval and highlight detection. $\otimes$ indicates element-wise multiplication.}
\label{fig2}
\end{figure*}

\section{Related Works}

\subsection{Moment Retrieval}
The connection between visual and language cues has become important in machine learning \cite{ lee2022audio, park2024enhancing, lee2024text}. Moment retrieval aims to locate relevant moments in a video based on a natural language query \cite{gao2017tall:charades-sta}. This task is typically approached using either proposal-based or proposal-free methods. The proposal-based methods \cite{gao2017tall:charades-sta, hendricks2018localizing:mr, xiao2021boundary:mr, sun2022you:mr} generate candidate proposals and rank them by matching scores. On the other hand, the proposal-free methods \cite{yuan2019find:mr_free, mun2020local:mr_free, rodriguez2020proposal:mr_free, li2021proposal:mr_free} directly regress the start and end timestamps through video-text interaction.

\subsection{Highlight Detection}
Highlight detection identifies the most significant parts of a video, which might not necessarily be tied to specific textual queries. Early highlight detection approaches are uni-modal, assessing the salience score of video clips without external textual data \cite{sun2014ranking:hd, xu2021cross:hd, wei2022learning:hd, badamdorj2022contrastive:hd}. However, as user preferences have increasingly influenced content consumption, integrating text queries has become common to tailor the detection process to individual user needs \cite{dagtas2004multimodal:hd_t, kudi2017words:hd_t, moment_detr2021:mr_hd}. It has been demonstrated that audio cues provide complementary information to visual features \cite{lee2022weakly, park2024learning, kim2024learning}. Advanced highlight detection systems now employ multi-modal inputs, including visual and audio cues, to enhance the detection accuracy and relevance of video highlights \cite{gygli2016video2gif:hd_a,  xiong2019less:hd_a, hong2020mini:hd_a, badamdorj2021joint:hd_a}.

Traditionally, moment retrieval and highlight detection are addressed separately, but recent work has explored their joint learning. MomentDETR \cite{moment_detr2021:mr_hd} introduces the QVHighlights dataset to facilitate joint learning of moment retrieval and highlight detection, proposing a DETR-based model. UMT \cite{umt2022:mr_hd} proposes adopting audio, visual, and text content for query generation to improve query quality. QD-DETR \cite{qd_detr2023:mr_hd} further leverages textual information by incorporating negative relationship learning between video-text pairs. UVCOM \cite{uvcom2024:mr_hd} and TR-DETR \cite{tr_detr2024:mr_hd} integrate the specialties of moment retrieval and highlight detection to achieve a comprehensive understanding.

Despite these advancements, many existing methods do not capture the overall context of the video. Capturing the overall context is essential for accurate moment retrieval and highlight detection as it provides a comprehensive understanding of the content and narrative flow. To this end, we present a video context-aware keyword attention module that understands the video context and identifies keywords between the video and text query.

\section{Proposed Method}

Figure \ref{fig2} shows the overall architecture of our framework. Similar to previous works \cite{qd_detr2023:mr_hd, tr_detr2024:mr_hd, uvcom2024:mr_hd}, we employ a two-stream network to extract video and text features. Input video of $L$ clips and a text query with $N$ words pass through each modal encoder (\textit{i.e.,} video encoder and text encoder) and three-layer feed-forward network to generate video features $\textbf{F}^v\in\mathbb{R}^{L\times d}$ and text features $\textbf{F}^t\in\mathbb{R}^{N\times d}$, respectively. To capture the overall flow of the video, $\textbf{F}^v$ is passed through a video context clustering module to generate $c$ clustered video features $\textbf{F}^{cv}\in\mathbb{R}^{c\times d}$. Subsequently, to identify video-related keywords in the text query, a keyword weight detection module calculates the similarity between clustered video features and text features, resulting in a keyword-weighted text feature $\textbf{F}^{wt}$. We then perform modality interaction between the video features and keyword-weighted text features, which is processed through a transformer encoder. Finally, we utilize the video-contextual information from $\textbf{F}^{cv}$ and cluster information $\textbf{C}$, along with the transformer-encoded feature, to conduct moment retrieval and highlight detection. More details are in the following subsections.

\subsection{Video Context-aware Keyword Attention Module}

To understand the entire video sequence and extract relevant text keywords corresponding to the video content, we propose a Video Context-aware Keyword Attention Module. 
As shown in Figure \ref{fig2}, our video context-aware keyword attention module consists of two steps: (1) video context clustering and (2) keyword weight detection.\\

\noindent{\textbf{(1) Video Context Clustering.}}
We propose a video context clustering module to cluster video clips, capturing the overall context of the video and identifying each scene. To account for the temporal order of the video, we adopt a temporally-weighted FINCH algorithm \cite{sarfraz2021temporally:clustering}. With video features $\textbf{F}^v\in\mathbb{R}^{L\times d}$ as input, the algorithm clusters video clips based on their adjacency relation and merges them hierarchically. Consequently, the most similar clips are grouped into $c$ clusters ($c_1,...c_c$) based on their relations. The output is a cluster information vector $\textbf{C}=\{C_{i}\}_{i=1}^{L}\in\mathbb{R}^{L}$, where each element $C_i \in \{c_1,...,c_c\}$ indicates the cluster assignment value for each video clip. Then, the clustered video features $\textbf{F}^{cv}$ are generated by averaging the information within each cluster.\\

\noindent{\textbf{(2) Keyword Weight Detection.}}
Aligning video and text features is essential for moment retrieval and highlight detection tasks. However, direct interaction between original video and text features can lead to information loss due to the misalignment \cite{xu2023multimodal:text_video_align}. Particularly in video moment retrieval and highlight detection tasks based on specific text queries, the emphasis on certain text keywords varies depending on the overall video context. Consider the text query \textit{``Chef makes pizza and cuts it up"} in Figure 2. From the perspective of moment retrieval and highlight detection, the word \textit{`chef'} might be less important if it appears consistently throughout the video. Instead, a sudden appearance of \textit{`pizza'} or the action of \textit{`cuts'} could be more significant. Conversely, if a machine is making pizza and a chef suddenly appears and makes pizza, \textit{`chef'} becomes a crucial keyword. As keyword importance changes with video content, we conduct keyword weight detection to identify the most important keyword related to the video.

To this end, we calculate cosine similarity matrix $M\in\mathbb{R}^{N\times c}$ between the text feature $\textbf{F}^t$ and the clustered video feature $\textbf{F}^{cv}$ (see Figure \ref{fig2}). We then apply each column-wise softmax and max-pooling to obtain a keyword weight vector $w^t\in\mathbb{R}^{N}$. Higher values in the keyword weight vector indicate words that are strongly associated with only specific clusters in the video, while lower values are associated with most clusters similarly.
Finally, we multiply $w^t$ with the original text feature $\textbf{F}^t$ to generate the keyword-weighted text feature $\textbf{F}^{wt}\in\mathbb{R}^{N\times d}$, which is be represented as:
\begin{gather}
M = \frac{\textbf{F}^t \, \textbf{F}^{cv\top}} {{|| \textbf{F}^t||}\,{||\textbf{F}^{cv}||}}, \\
w^t = \text{MaxPooling}(\text{Softmax}(M / \tau)),\\
\textbf{F}^{wt} = w^t \textbf{F}^{t},
\end{gather}
where $\tau$ is a temperature hyper-parameter. This keyword-weighted text feature $\textbf{F}^{wt}$ emphasizes contextually important words in the query, enhancing video-text alignment and improving performance in moment retrieval and highlight detection tasks.

\subsection{Video Contextual MR/HD Prediction}
Moment retrieval (MR) and highlight detection (HD) are two crucial tasks in video understanding. Moment retrieval aims to localize the center coordinates and duration of moments related to a given text query, while highlight detection generates a saliency score distribution across the entire video.  To improve the effectiveness of these tasks, we utilize the video-contextual information $\textbf{C}$ and  $\textbf{F}^{cv}$ generated by the video context clustering module in each prediction head.

In the context of moment retrieval, transition points between clusters are crucial. These points often correspond to scene changes in the video, providing valuable information for the moment retrieval task. By leveraging the cluster information vector $\textbf{C}$, we create a binary context change vector $\textbf{C}^m\in\mathbb{R}^{L}$ that encapsulates information about these transitions. In this vector, we assign a value of 1 if the cluster number changes between the $i$-th and ($i$+1)-th frame (\textit{e.g.,} $c_j$$\rightarrow$$c_{j+1}$), and 0 otherwise. Then, following previous works \cite{qd_detr2023:mr_hd}, we employ a standard transformer decoder structure for moment retrieval head. However, instead of the traditional approach, we use $\textbf{C}^m$ as the initial embedding for the learnable anchors, which helps the MR decoder better focus on scene transition points in the video, potentially leading to more accurate moment identification.

For highlight detection, we focus on the representative values of each cluster obtained through our clustering approach. These values provide information about the best representation of each scene context. To obtain this information, we compute a cosine similarity between each video clip feature $\textbf{F}^v$ and the average information of the cluster it belongs to, which we extract from $\textbf{F}^{cv}$ using $\textbf{C}$. This similarity computation results in $\textbf{C}^h\in\mathbb{R}^{L}$, which indicates how well each clip represents the information of its cluster. To generate saliency score distribution for highlight detection, we use two groups of single fully connected layers for linear projection to calculate the saliency score, following \cite{qd_detr2023:mr_hd}. As input to this process, we use a context-aware video token $T^{cv} \in \mathbb{R}^{L \times (d+1)}$, which is created by concatenating $\textbf{C}^h$ with the video token $T^v\in\mathbb{R}^{L \times d}$ obtained from passing through a transformer encoder. The predicted saliency scores $\textbf{S}$ are then computed using the following equation:
\begin{equation}
\begin{gathered}
\textbf{S} = \frac{T^s w^{s\top} \cdot T^{cv}w^{cv\top}}{\sqrt{p}}, \
\end{gathered}
\end{equation}
where $T^s\in\mathbb{R}^{d}$ is the randomly initialized input-adaptive saliency token, $w^s\in\mathbb{R}^{p\times d}$ and $w^{cv}\in\mathbb{R}^{p \times{(d+1)}}$ are learnable parameters, and $p$ is the projection dimension.

\subsection{Keyword-aware Contrastive Loss}
To enhance the alignment between text query features and video features by leveraging the overall flow of the video, we introduce keyword-aware contrastive loss. This loss is composed of two components: a clip-keyword contrastive loss and a video-keyword contrastive loss. The clip-keyword contrastive loss focuses on intra-video relationships between text queries and visual features of each clip, while the video-keyword contrastive loss addresses inter-video relationships across the dataset.\\

\noindent{\textbf{Clip-keyword Contrastive Loss.}}
Existing methods \cite{uvcom2024:mr_hd, tr_detr2024:mr_hd} typically construct loss functions that bring the clip features of ground-truth moments closer to the text query features while pushing background clip features away. However, as illustrated in Figure \ref{fig1}, even background clips considered irrelevant to the text query may still have high relevance to specific words in the text query. In such cases, the feature of background clips can be misrepresented through contrastive loss from existing methods. To address this, we utilize the keyword weight vector $w^t$ to emphasize keywords in advance. This approach enables more robust alignment between the clip features $\textbf{F}^v$ and the keyword-weighted text features $\textbf{F}^{wt} = w^t\textbf{F}^t$ of Eq.(3). We formulate the clip-keyword contrastive loss $\mathcal{L}_{ck}$ as follows:

\begin{table*}[t]
    \renewcommand{\tabcolsep}{3.3mm}
    \centering
	\resizebox{0.97\linewidth}{!}{
		\begin{tabular}{ccccccccc}
            \Xhline{3\arrayrulewidth}
            \rule{0pt}{10.0pt} \multirow{3}{*}[-0.35em]{\bf Method} & \multirow{3}{*}[-0.35em]{Src.} & \multicolumn{5}{c}{\textbf{MR} } & \multicolumn{2}{c}{\textbf{HD} } \\ 
            \cmidrule(lr){3-7} \cmidrule(l){8-9}
            && \multicolumn{2}{c}{R1} & \multicolumn{3}{c}{mAP} & \multicolumn{2}{c}{$\geq$Very Good} \\
            \cmidrule(lr){3-4} \cmidrule(lr){5-7} \cmidrule(l){8-9} 
            && @0.5 & @0.7 & @0.5 & @0.75 & Avg. & mAP & HIT@1 \\
            \midrule
            M-DETR \cite{moment_detr2021:mr_hd} & $\mathcal{V}$ & 52.89 & 33.02 & 54.82 & 29.40 & 30.73 & 35.69 & 55.60 \\
            QD-DETR \cite{qd_detr2023:mr_hd} & $\mathcal{V}$ & 62.40 & 44.98 & 62.52 & 39.88 & 39.86 & 38.94 & 62.40 \\
            UniVTG \cite{lin2023univtg:uni-vtg} & $\mathcal{V}$ & 58.86 & 40.86 & 57.60 & 35.59 & 35.47 & 38.20 & 60.96 \\
            TR-DETR \cite{tr_detr2024:mr_hd} & $\mathcal{V}$ & \underline{64.66} & \underline{48.96} & \underline{63.98} & \underline{43.73} & 42.62 & \underline{39.91} & 63.42 \\
            UVCOM \cite{uvcom2024:mr_hd} & $\mathcal{V}$ & 63.55 & 47.47 & 63.37 & 42.67 & \underline{43.18} & 39.74 & \underline{64.20} \\\hdashline
            \rule{0pt}{9.5pt}
            \textbf{Ours} & $\mathcal{V}$ & \textbf{66.86} & \textbf{51.23} & \textbf{67.73} & \textbf{46.24} & \textbf{45.69} & \textbf{40.94} & \textbf{64.79} \\
            \hline
            \rule{0pt}{9.5pt}
            UMT \cite{umt2022:mr_hd} & $\mathcal{V}+\mathcal{A}$ & 56.23 & 41.18 & 53.38 & 37.01 & 36.12 & 38.18 & 59.99 \\
            QD-DETR \cite{qd_detr2023:mr_hd} & $\mathcal{V}+\mathcal{A}$ & 63.06 & 45.10 & 63.04 & 40.10 & 40.19 & 39.04 & 62.87 \\
            TR-DETR \cite{tr_detr2024:mr_hd} & $\mathcal{V}+\mathcal{A}$ & \underline{65.05} & 47.67 & \underline{64.87} & 42.98 & 43.10 & \underline{39.90} & 63.88 \\
            UVCOM \cite{uvcom2024:mr_hd} & $\mathcal{V}+\mathcal{A}$ & 63.81 & \underline{48.70} & 64.47 & \underline{44.01} & \underline{43.27} & 39.79 & \underline{64.79} \\\hdashline
            \rule{0pt}{9.5pt}
            \textbf{Ours} & $\mathcal{V}+\mathcal{A}$ & \textbf{67.77} & \textbf{50.52} & \textbf{68.30} & \textbf{45.88} & \textbf{45.52} & \textbf{41.15} & \textbf{65.82} \\
            \Xhline{3\arrayrulewidth}
            \end{tabular}
        }
    \caption{Experimental results on the QVHighlights \textit{test} set for moment retrieval and highlight detection when using either video only ($\mathcal{V}$) or video and audio ($\mathcal{V}+\mathcal{A}$). \textbf{Bold}/\underline{underlined} fonts indicate the best/second-best results.}
    \label{qvhighlight}
\end{table*}

\begin{gather}
\textbf{G}^{wt} = \text{MeanPooling}(\textbf{F}_i^{wt}), \\
\mathcal{L}_{ck} = -\frac{1}{B} \sum_{i=1}^{B} \log \frac{\sum_{j\in R_i} exp(Sim(\textbf{F}^{v_{(j)}}_i, \textbf{G}^{wt}_i))}{\sum_{j=1}^L exp(Sim(\textbf{F}^{v_{(j)}}_i, \textbf{G}^{wt}_i))},\\
Sim(A,B) =\frac{AB^{\top}} {{||A||}\,{||B||}},
\end{gather}\\
where $\textbf{G}_i^{wt}\in\mathbb{R}^{d}$ is average of word-level text feature, $R_i$ denotes relevant ground-truth clips in the \textit{i}-th video and $B$ indicates the batch number. The $\mathcal{L}_{ck}$ maximizes the learning effect of essential central information, thereby enabling more accurate moment retrieval and highlight detection.\\

\noindent{\textbf{Video-keyword Contrastive Loss.}}
Extending beyond single video contexts, we propose a global contrastive loss, called video-keyword contrastive loss, to operate across the entire dataset. Unlike existing methods \cite{uvcom2024:mr_hd,tr_detr2024:mr_hd} that use unweighted global information from video-text pairs, we incorporate keyword-weighted text features $\textbf{F}^{wt}$ to obtain a better global representation by utilizing the global information of relevant videos and keyword-weighted text queries. We define the video-keyword contrastive loss $\mathcal{L}_{vk}$ as:
\begin{gather}
\textbf{G}^{v}_i = \text{MeanPooling}(r^b_i\textbf{F}^v_i), \\
\mathcal{L}_{vk} = -\frac{1}{B} \sum_{i=1}^{B} \log \frac{exp(Sim(\textbf{G}^{v}_i, \textbf{G}^{wt}_i))}{\sum_{j=1}^B exp(Sim(\textbf{G}^{v}_j, \textbf{G}^{wt}_i))},
\end{gather}
where $\textbf{G}^{v}\in\mathbb{R}^{d}$ is average of clip-level visual feature, $r^b$ is a binary value (1 for ground-truth clips, 0 otherwise). The $\mathcal{L}_{vk}$ strengthens the global representation based on keywords, facilitating more effective cross-video learning.

Finally, we devise a keyword-aware contrastive loss $\mathcal{L}_{kw}$ that combines $\mathcal{L}_{ck}$ and $\mathcal{L}_{vk}$, which can be formulated as:
\begin{equation}
\mathcal{L}_{kw} = \mathcal{L}_{ck} + \mathcal{L}_{vk}.
\end{equation}
The $\mathcal{L}_{vk}$ enables our model to optimize both temporal relevance within videos and global semantic coherence across the dataset, achieving a comprehensive alignment between text queries and video contents.

\subsection{Training Objective}
To train our proposed method, we construct the total training loss function as follows:
\begin{equation}
\mathcal{L}_{Total}=\mathcal{L}_{mr}+\mathcal{L}_{hd}+\lambda_{kw}\mathcal{L}_{kw},
\end{equation}
where $\mathcal{L}_{mr}$ and $\mathcal{L}_{hd}$ denote the loss functions for moment retrieval and highlight detection as outlined in \cite{qd_detr2023:mr_hd}. $\lambda_{kw}$ is a balancing parameter. The $\mathcal{L}_{Total}$ enables effective moment retrieval and highlight detection.

\begin{table*}[t!]
    \renewcommand{\tabcolsep}{3.0mm}
    \centering
	\resizebox{0.97\linewidth}{!}{
		\begin{tabular}{cccccccccccc}
            \Xhline{3\arrayrulewidth}
            \rule{0pt}{10.0pt}
            \textbf{Method} & VT & VU & GA & MS & PK & PR & FM & BK & BT & DS & Avg. \\
            \hline
            \rule{0pt}{9.5pt}
            sLSTM \cite{zhang2016video:slstm} & 41.1 & 46.2 & 46.3 & 47.7 & 44.8 & 46.1 & 45.2 & 40.6 & 47.1 & 45.5 & 45.1 \\
            LIM-S \cite{xiong2019less:hd_a} & 55.9 & 42.9 & 61.2 & 54.0 & 60.3 & 47.5 & 43.2 & 66.3 & 69.1 & 62.6 & 56.3 \\
            Trailer \cite{wang2020learning:trailer} & 61.3 & 54.6 & 65.7 & 60.8 & 59.1 & 70.1 & 58.2 & 64.7 & 65.6 & 68.1 & 62.8 \\
            SL-Module \cite{xu2021cross:hd} & 86.5 & 68.7 & 74.9 & \underline{86.2} & 79.0 & 63.2 & 58.9 & 72.6 & 78.9 & 64.0 & 73.3 \\
            UMT† \cite{umt2022:mr_hd} & 87.5 & 81.5 & 88.2 & 78.8 & 81.5 & 87.0 & 76.0 & 86.9 & 84.4 & 79.6 & 83.1 \\
            QD-DETR \cite{qd_detr2023:mr_hd} & 88.2 & 87.4 & 85.6 & 85.0 & 85.8 & 86.9 & 76.4 & 91.3 & 89.2 & 73.7 & 85.0 \\
            UniVTG \cite{lin2023univtg:uni-vtg} & 83.9 & 85.1 & 89.0 & 80.1 & 84.6 & 87.0 & 70.9 & 91.7 & 73.5 & 69.3 & 81.0 \\
            TR-DETR \cite{tr_detr2024:mr_hd} & \underline{89.3} & \underline{93.0} & \underline{94.3} & 85.1 & \underline{88.0} & \underline{88.6} & \underline{80.4} & 91.3 & \underline{89.5} & \textbf{81.6} & \underline{88.1} \\
            UVCOM \cite{uvcom2024:mr_hd} & 87.6 & 91.6 & 91.4 & \textbf{86.7} & 86.9 & 86.9 & 76.9 & \underline{92.3} & 87.4 & 75.6 & 86.3 \\\cdashline{1-12}
            \rule{0pt}{10.2pt}
            \textbf{Ours} & \textbf{89.9} & \textbf{93.8} & \textbf{94.4} &
            85.9 & \textbf{89.2} & \textbf{89.4} & \textbf{81.5} & \textbf{92.6} & \textbf{90.1} & \underline{80.6} & \textbf{88.7} \\
            \Xhline{3\arrayrulewidth}
            \end{tabular}
        }
    \caption{Experimental results on the TVSum for highlight detection. † indicates training with audio modality. \textbf{Bold}/\underline{underlined} fonts indicate the best/second-best results.}
    \label{tvsum}
\end{table*}

\section{Experiments}
\subsection{Datasets and Evaluation Metrics}
\noindent{\textbf{QVHighlights.}} 
The QVHighlights dataset \cite{moment_detr2021:mr_hd} includes 10,148 YouTube videos with rich content, each paired with an annotated text query that indicates highlight moments. This is the only dataset that includes both annotations for moment retrieval and highlight detection. Following \cite{moment_detr2021:mr_hd}, to ensure a fair evaluation, we submitted our model predictions to the QVHighlights server CodaLab competition platform, with test set annotations remaining confidential. \\

\noindent{\textbf{TVSum.}}
The TVSum dataset \cite{song2015tvsum:tvsum} is also a standard benchmark for highlight detection, comprising videos from 10 different categories, with each category containing 5 videos. For a fair comparison, we use the same train/test split as utilized in QD-DETR\cite{qd_detr2023:mr_hd}.\\

\noindent{\textbf{Charades-STA.}}
The Charades-STA dataset \cite{gao2017tall:charades-sta} contains 9,848 videos depicting indoor activities with 16,128 human-annotated query texts. Following QD-DETR \cite{qd_detr2023:mr_hd}, we use 12,408 samples for training, with the remaining 3,720 samples allocated for testing. \\

\begin{table}[t!]
    \renewcommand{\tabcolsep}{1.6mm}
    \centering
	\resizebox{0.97\linewidth}{!}{
		\begin{tabular}{cccc}
            \Xhline{3\arrayrulewidth}
            \rule{0pt}{10.0pt}
            \textbf{Method} & \textbf{Feat} 
            & R1@0.5 & R1@0.7 \\
            \hline
            \rule{0pt}{9.5pt}
            2D-TAN \cite{zhang20202dtan:moment_retrieval} & VGG & 40.94 & 22.85 \\
            UMT† \cite{umt2022:mr_hd} & VGG & 48.31 & 29.25 \\
            QD-DETR \cite{qd_detr2023:mr_hd} & VGG & 52.77 & \underline{31.13} \\
            TR-DETR \cite{tr_detr2024:mr_hd} & VGG & \underline{53.47} & 30.81 \\
            \cdashline{1-4}
            \rule{0pt}{9.5pt}
            \textbf{Ours} & VGG & \textbf{54.89} & \textbf{31.97} \\
            \hline
            \rule{0pt}{9.5pt}
            QD-DETR \cite{qd_detr2023:mr_hd} & SF+C & 57.31 & 32.55 \\
            UniVTG \cite{lin2023univtg:uni-vtg} & SF+C & 58.01 & 35.65 \\
            TR-DETR \cite{tr_detr2024:mr_hd} & SF+C & 57.61 & 33.52 \\
            UVCOM \cite{uvcom2024:mr_hd} & SF+C & \underline{59.25} & \underline{36.64} \\\cdashline{1-4}
            \rule{0pt}{9.5pt}
            \textbf{Ours} & SF+C & \textbf{61.08} & \textbf{37.89} \\
            \Xhline{3\arrayrulewidth}
            \end{tabular}
        }
    \caption{Experimental results on the Charades-STA for moment retrieval. † indicates training with audio modality. \textbf{Bold}/\underline{underlined} fonts indicate the best/second-best results. }
    \label{charades}
\end{table}

\noindent{\textbf{Evaluation Metric.}}
For the evaluation, we follow the metrics of prior works \cite{moment_detr2021:mr_hd, uvcom2024:mr_hd, tr_detr2024:mr_hd} for a fair and comprehensive comparison. In the QVHighlights dataset, we evaluate Recall@1 (R1) at IoU thresholds of 0.5 and 0.7, and mean average precision (mAP) at thresholds from 0.5 to 0.95 in steps of 0.05 (mAP@Avg). We compare performance at IoU thresholds of 0.5 and 0.75, referred to as mAP@0.5 and mAP@0.75. For highlight detection, we use mAP and HIT@1 (hit ratio of the highest-scored clip). In the Charades-STA dataset, we evaluate Recall@1 at IoU thresholds of 0.5 and 0.7. For the TVSum dataset, the primary evaluation metric is top-5 mAP.

\subsection{Implementation Details}

\noindent{\textbf{Pre-extracted Features.}} Following \cite{qd_detr2023:mr_hd}, we use the pre-extracted video, text, and audio features from the various models. For video features, we use the pre-trained SlowFast \cite{feichtenhofer2019slowfast:slowfast} and CLIP \cite{radford2021learning:clip} models for QVHighlights, VGG \cite{simonyan2014very:vgg} and SlowFast+CLIP (SF+C) for Charades-STA, and I3D pre-trained on Kinetics 400 \cite{carreira2017quo:kinetic_i3d} for TVSum. For text features, we use CLIP \cite{radford2021learning:clip} for QVHighlights and TVSum, and GloVe \cite{pennington2014glove} for Charades-STA. We use audio features from all datasets using a PANN \cite{kong2020panns:pann} model pre-trained on AudioSet. \\

\noindent{\textbf{Training Settings.}} 
We set the loss weights to $\lambda_{kw}=0.3$ and use the Adam optimizer \cite{kingma2014adam} with a learning rate of 1e-3 and a weight decay of 1e-4. We train QVHighlights with a batch size of 32 for 200 epochs, Charades-STA with a batch size of 8 for 100 epochs, and TVSum with a batch size of 4 for 2000 epochs, using a single RTX 4090 GPU. For detailed network configurations, please refer to the supplementary material.

\begin{table}[t]
    \renewcommand{\tabcolsep}{2.3mm}
    \centering
	\resizebox{0.99\linewidth}{!}{
		\begin{tabular}{ccccccc}
            \Xhline{3\arrayrulewidth}
            \rule{0pt}{10.0pt}
            \multirow{3}{*}[-0.8ex]{\bf VCKA} & \multirow{3}{*}[-0.8ex]{\bf VCP} & \multicolumn{3}{c}{\textbf{MR} } & \multicolumn{2}{c}{\textbf{HD} } \\
            \cmidrule(lr){3-5} \cmidrule(l){6-7}
            && \multicolumn{2}{c}{R1} & \multirow{2}{*}[-0.6ex]{\makecell{mAP\\Avg.}} & \multirow{2}{*}[-0.6ex]{mAP} & \multirow{2}{*}[-0.6ex]{HIT@1} \\\cmidrule(l){3-4}
            & & @0.5 & @0.7 & & \\
            \midrule
            - & - & 66.39 & 49.03 & 45.55 & 40.78 & 65.42 \\\cdashline{1-7}
            \rule{0pt}{10.0pt}
            \cmark & - & 68.90 & 52.32 & 46.98 & 41.38 & 66.19 \\
            - & \cmark& 66.90 & 51.03 & 46.42 & 41.30 & 66.71 \\
            \cmark & \cmark & \textbf{68.97} & \textbf{53.35} & \textbf{47.69} & \textbf{41.67} & \textbf{67.03} \\
            \Xhline{3\arrayrulewidth}
            \end{tabular}
        }
    \caption{Effect of the proposed component (video context-aware keyword attention (VCKA) module, and video contextual MR/HD prediction (VCP)) on QVHighlights \textit{val} set.}
    \label{component_ablation}
\end{table}

\begin{figure*}[t]
\centering
    \includegraphics[width=0.97\textwidth]{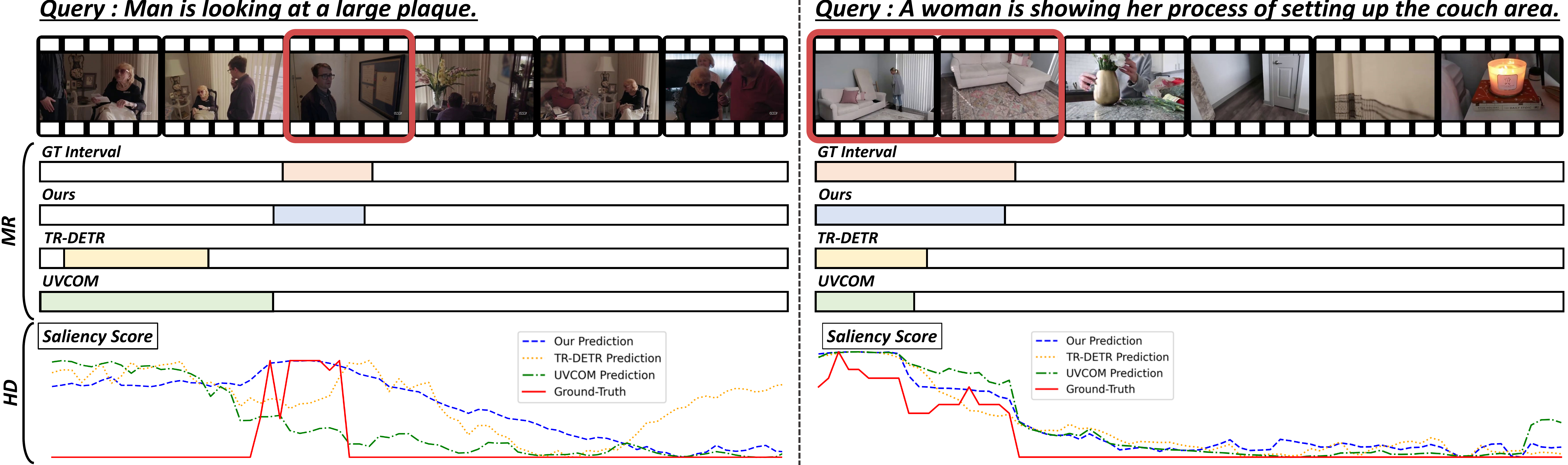}
\caption{Visualization examples for moment retrieval (MR) and highlight detection (HD) on the QVHighlights \textit{val} set.}
\label{fig3}
\end{figure*}

\subsection{Comparison to Prior Works}
\noindent{\textbf{Results on the QVHighlights.}}
Table 1 compares the performance of our method with existing state-of-the-art methods on the QVHighlights dataset for moment retrieval (MR) and highlight detection (HD). When using only the video ($\mathcal{V}$), our method shows superior performance, achieving an average of 2.12\% higher performance across all metrics. A similar trend is observed when combining video with audio ($\mathcal{V}+\mathcal{A}$). This indicates that the keyword-aware contrastive loss $\mathcal{L}_{kw}$ effectively enables our framework to understand video content and capture relevant keywords. \\

\noindent{\textbf{Results on the TVSum.}}
We use the TVSum dataset to evaluate video highlight detection performance, following protocols of previous works \cite{qd_detr2023:mr_hd, tr_detr2024:mr_hd, uvcom2024:mr_hd}. As shown in Table 2, our method shows superior performance in 8 out of 10 categories, with an overall average performance (Avg.) of 88.7\%, indicating an improvement over existing methods. \\

\noindent{\textbf{Results on the Charades-STA.}}
We also evaluate the performance of our method on moment retrieval using the Charades-STA dataset. As shown in Table 3, our approach consistently shows improved performance, achieving a 0.84\% improvement over VGG feature and a 1.25\% improvement over SF+C (SlowFast+CLIP) feature in R1@0.7. This demonstrates that our method is robust across various pre-extracted video features for the moment retrieval.

\subsection{Ablation Study}
In this section, we conduct various ablation studies to investigate (1) the effect of our proposed components and (2) the effect of our proposed losses. All ablation studies are conducted using QVHighlights \textit{val} set to evaluate both moment retrieval and highlight detection.\\

\noindent{\textbf{Effect of the Proposed Components.}}
Table 4 shows the effect of the two components Video Context-aware Keyword Attention module (VCKA) and Video Contextual MR/HD Prediction (VCP) heads. The results clearly demonstrate that both of proposed components bring better performance, highlighting the effectiveness of our approach. \\

\begin{table}[t]
    \renewcommand{\tabcolsep}{2.3mm}
    \centering
	\resizebox{0.99\linewidth}{!}{
		\begin{tabular}{cc ccccc}
            \Xhline{3\arrayrulewidth}
            \rule{0pt}{10.0pt}
            \multirow{3}{*}[-0.8ex]{\bf $\mathcal{L}_{ck}$} & \multirow{3}{*}[-0.8ex]{\bf $\mathcal{L}_{vk}$} & \multicolumn{3}{c}{\textbf{MR} } & \multicolumn{2}{c}{\textbf{HD} } \\
            \cmidrule(lr){3-5} \cmidrule(l){6-7}
            & & \multicolumn{2}{c}{R1} & \multirow{2}{*}[-0.6ex]{\makecell{mAP\\Avg.}} & \multirow{2}{*}[-0.6ex]{mAP} & \multirow{2}{*}[-0.6ex]{HIT@1} \\\cmidrule(l){3-4}
            & & @0.5 & @0.7 & & \\
            \midrule
            - &- & 66.90 & 49.94 & 45.02 & 40.53 & 65.48 \\\cdashline{1-7}
            \rule{0pt}{9.5pt}
            \cmark& - & 67.74 & 51.16 & 46.37 & 41.32 & 66.45 \\
            - & \cmark & 67.16 & 51.42 & 46.76 & 40.84 & 65.10 \\
            \cmark & \cmark & \textbf{68.97} & \textbf{53.35} & \textbf{47.69} & \textbf{41.67} & \textbf{67.03} \\
            \Xhline{3\arrayrulewidth}
            \end{tabular}
        }
    \caption{Effect of the proposed keyword-aware contrastive losses on QVHighlights \textit{val} set.}
    \label{loss_ablation}
\end{table}

\noindent{\textbf{Proposed Losses.}}
We evaluate our method with respect to the two keyword-aware contrastive losses, $\mathcal{L}_{ck}$ and $\mathcal{L}_{vk}$. The results in Table 5 show that considering each loss individually consistently surpasses the baseline, which does not use these contrastive losses. Combining all proposed losses achieves the highest performance, significantly enhancing the method by improving its capacity to learn robust and discriminative video features.

\subsection{Visualization Results}

As shown in Figure 3, We provide examples of moment retrieval and highlight detection on QVHighlights \textit{val} set by comparing our method with TR-DETR and UVCOM. The visualizations demonstrate the efficacy of our method in both moment retrieval and highlight detection.
\begin{figure}[t]
\begin{minipage}[b]{1.0\linewidth}
\centering
\centerline{\includegraphics[width=0.995\linewidth]{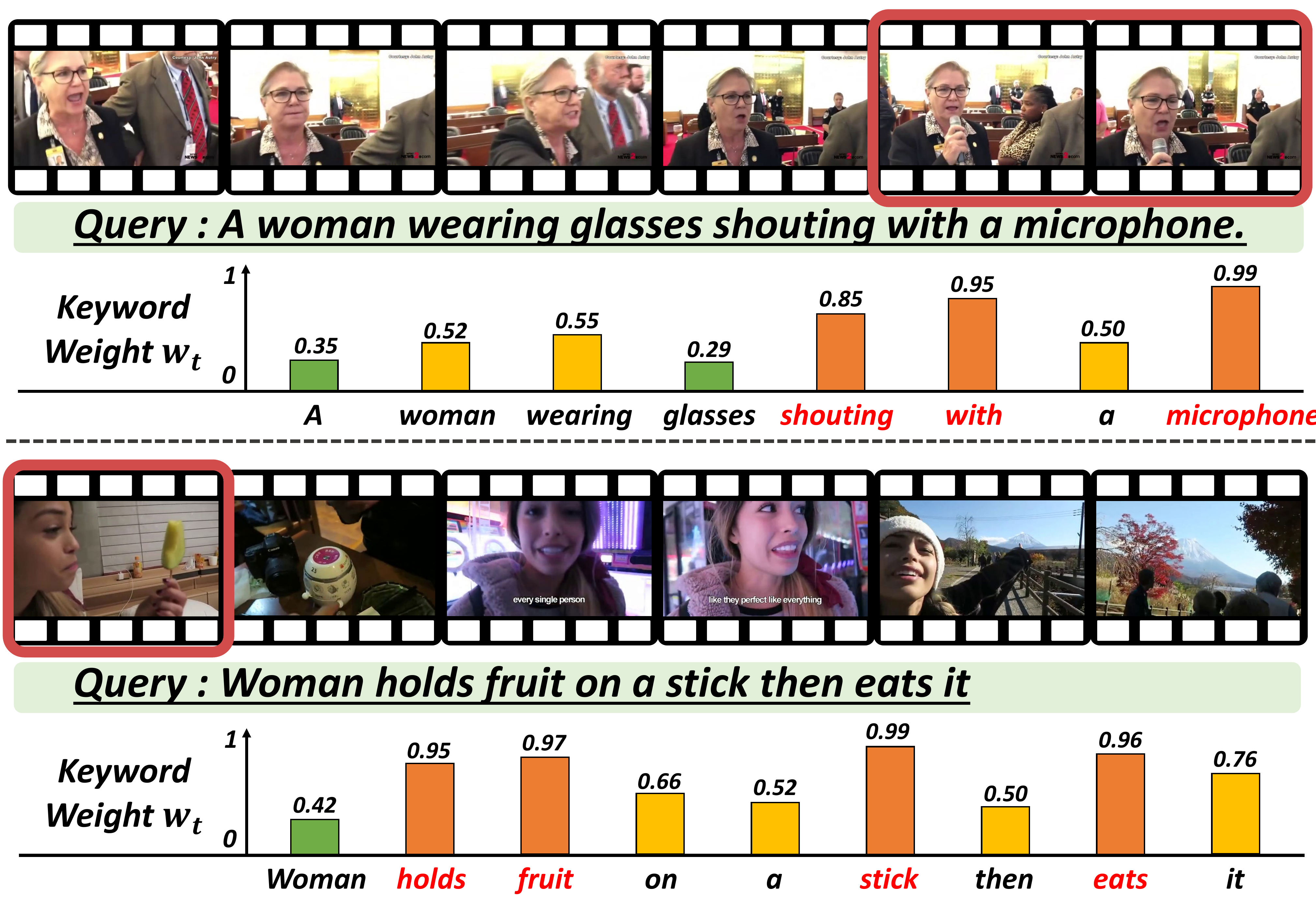}}
\end{minipage}
\caption{Visualization results of keyword weight effectiveness on QVHighlights \textit{val} set.}
\label{fig4}
\end{figure}

\subsection{Discussion}

\noindent{\textbf{Effectiveness of Our Keyword Weight.}}
We visualize the keyword weight $w^t$ of each query word on QVHighlights \textit{val} set. As shown in Figure 4, in the first sample, \textit{`woman wearing glasses'} appears in most scenes, resulting in low keyword weights. Conversely, \textit{`microphone'} receives a high weight due to its specificity in the video context. Similarly, in the second sample, \textit{`woman'} has a low weight due to frequent appearance, while \textit{`fruit'}, \textit{`stick'}, and \textit{`eats'} get higher weights. These visualizations demonstrate the effectiveness of our keyword weight $w^t$ which can capture contextual importance within video. \\

\subsection{Limitation}
Our method has demonstrated robust performance across diverse datasets. However, when processing both visual and audio inputs, we use a simplified audio representation without a detailed audio-specific framework. Therefore, developing a more sophisticated audio-visual integration approach is an important direction for future work.

\section{Conclusion}
In this paper, we propose an innovative approach to moment retrieval and highlight detection that improves understanding of the video context and utilizes contextually appropriate keywords. The core of our method is the video context-aware keyword attention module, which effectively comprehends the entire video sequence and extracts relevant text keywords corresponding to the video content. Our keyword-aware contrastive loss functions successfully enhance the alignment between text query features and video clip features by leveraging the overall flow of the video. We believe that our method not only enhances accuracy but also holds diverse practical applications.

\section{Acknowledgments}
This work was supported by the NRF grant funded by the Korea government (MSIT) (No. RS-2023-00252391), and by the IITP grant funded by the Korea government (MSIT) (No. RS-2022-00155911: Artificial Intelligence Convergence Innovation Human Resources Development (Kyung Hee University), No. 2022-0-00124: Development of Artificial Intelligence Technology for Self-Improving Competency-Aware Learning Capabilities, IITP-2023-RS-2023-00266615: Convergence Security Core Talent Training Business Support Program), and conducted by CARAI grant funded by DAPA and ADD (UD230017TD).

\bibliography{aaai25}

\setcounter{section}{0}
\setcounter{figure}{0}
\setcounter{table}{0}
\setcounter{equation}{0}
\pagenumbering{gobble}

\maketitlesupplementary

\noindent This manuscript provides additional implementation details, ablation studies, and visualization comparisons that further support the main paper.

\section{Additional Implementation Details}

\noindent{\textbf{Network Configuration Details.}} We use 3 decoder transformer layers for the QVHighlights and Charades-STA datasets, and 1 for TVSum, which is smaller in scale. The number of bottleneck tokens is fixed at 4, with hidden dimensions of 256 and a 4x expansion in feed-forward networks. We apply learnable positional encodings, pre-norm style layer normalizations, eight attention heads, and a uniform dropout rate of 0.1 across all transformer layers. For the TVSum dataset, we utilize clips with a saliency score of 3 or higher as relevant ground-truth clips for our keyword-aware contrastive loss. We also implement additional pre-dropout rates: 0.5 for visual and audio inputs, and 0.3 for text inputs. All experiments are conducted using PyTorch, leveraging the codebase of Moment-DETR \cite{moment_detr2021:mr_hd}.

\section{Additional Experiments}

\noindent{\textbf{Variation of $\lambda_{kw}$.}}
To evaluate the effect of the balancing parameter $\lambda_{kw}$ in our keyword-aware contrastive loss $\mathcal{L}_{kw}$, we perform an additional study. As shown in Table \ref{loss_ablation}, our model achieves optimal performance when $\lambda_{kw}$ is set to 0.3. Remarkably, our method consistently surpasses the existing approaches in main paper across a range of $\lambda_{kw}$ values. This indicates that while there is an optimal setting that maximizes the effectiveness of our model, the model exhibits robustness against variations in this parameter. This robustness underscores the stability and efficacy of our proposed method, even when the parameter is not optimally tuned. \\

\noindent{\textbf{Computational Costs.}}
Table \ref{table:cost} compares training and inference times, and the number of parameters, of our method with TR-DETR and UVCOM, which exhibit the top two performances among existing methods. Our method shows a slight increase in training time compared to TR-DETR due to the clustering process but remains more efficient than UVCOM. Note that, with CLIP feature extraction included, it only incurs a minor (2\%) increase in inference time relative to TR-DETR, demonstrating its competitive efficiency.

\section{Additional Visualization Results}
We provide further comparisons of visualization results between our method and TR-DETR \cite{tr_detr2024:mr_hd} and UVCOM \cite{uvcom2024:mr_hd}, which exhibit the best and second-best performance among existing methods on the QVHighlights dataset (QVHighlights \textit{val}). It is shown in Figure 1. Our keyword-aware method outperforms others in both moment retrieval (MR) and highlight detection (HD) tasks. The additional visualization results further illustrate the superior performance of our method.

\begin{table}[t]
    \renewcommand{\tabcolsep}{3.2mm}
    \centering
		\resizebox{0.99\linewidth}{!}{
		\begin{tabular}{cccccc}
            \Xhline{3\arrayrulewidth}
            \rule{0pt}{10.0pt}
            \multirow{3}{*}[-0.8ex]{\bf $\lambda_{kw}$} & \multicolumn{3}{c}{\textbf{MR} } & \multicolumn{2}{c}{\textbf{HD} } \\
            \cmidrule(lr){2-4} \cmidrule(l){5-6}
            & \multicolumn{2}{c}{R1} & \multirow{2}{*}[-0.6ex]{\makecell{mAP\\Avg.}} & \multirow{2}{*}[-0.6ex]{mAP} & \multirow{2}{*}[-0.6ex]{HIT@1} \\\cmidrule(l){2-3}
            & @0.5 & @0.7 & & \\
            \midrule
            0.1 & 67.94 & 52.13 & 46.65 & 41.49 & 66.90 \\
            \textbf{0.3} & \underline{68.97} & \textbf{53.35} & \textbf{47.69} & \textbf{41.67} & \textbf{67.03} \\
            0.5 & 68.71 & \underline{52.90} & \underline{47.68} & 41.58 & 66.97 \\
            0.7 & 68.53 & 52.67 & 46.89 & 41.52 & 66.04 \\
            0.9 & \textbf{69.29} & 52.52 & 47.57 & \underline{41.62} & \underline{67.01} \\
            \Xhline{3\arrayrulewidth}
            \end{tabular}
        }
    \caption{Additional experimental results according to the balancing parameter $\lambda_{kw}$ on the QVHighlights \textit{val} set. \textbf{Bold}/\underline{underlined} fonts indicate the best/second-best results.}
    \label{loss_ablation}
\end{table}

		\begin{table}[t]
			\renewcommand{\tabcolsep}{0.4mm}
			\centering
			\resizebox{1.0\linewidth}{!}{
				\begin{tabular}{cccc}
					\Xhline{3\arrayrulewidth}
					\rule{0pt}{9.5pt} \bf \multirow{2}{*}{\bf Method} & \bf Train. (s) & \bf Infer. (s) & \multirow{2}{*}{\bf \#params}\\
					& \bf (\textit{per iter}) & \bf (\textit{per video})
					\\ \hline
					\rule{0pt}{10.pt}
                    CLIP F.E. & - & 0.140 & - \\\hdashline
                    \rule{0pt}{9.5pt}
                    
					TR-DETR \cite{tr_detr2024:mr_hd}      & 0.026 & 0.010 & 8.30M  \\
                    UVCOM \cite{uvcom2024:mr_hd} & 0.059 & 0.022 & 17.59M \\
					\textbf{Ours} & 0.038 & 0.013 & 8.32M \\
					\Xhline{3\arrayrulewidth}
				\end{tabular}
			}
            \caption{The comparisons of training time, inference time, and the number of parameters. CLIP F.E. denotes CLIP feature extraction, which is mandatory for all methods.}

			\label{table:cost}
		\end{table}

\begin{figure*}[t]
\centering
    \includegraphics[width=0.99\textwidth]{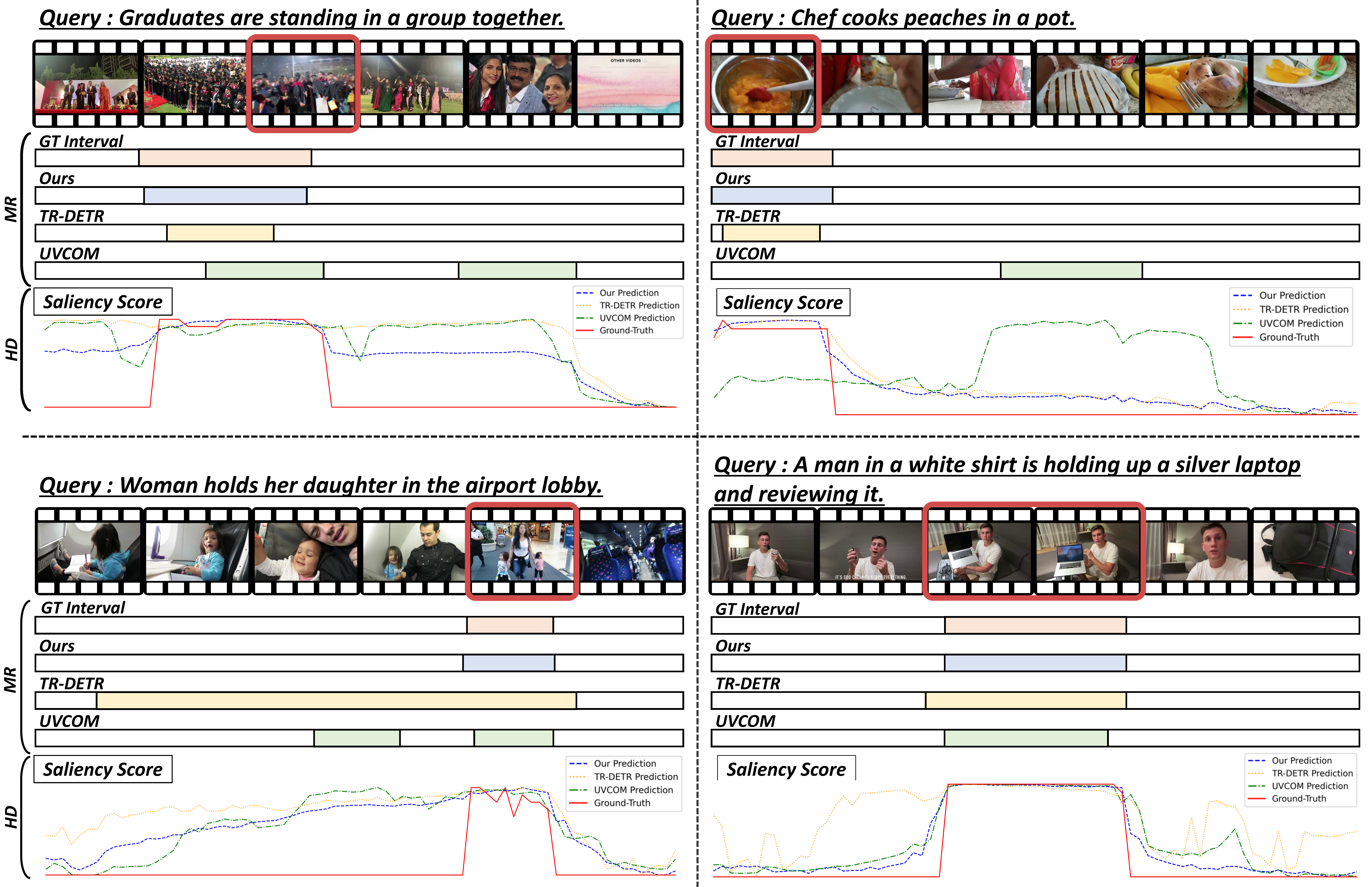}
\caption{Additional visualization comparisons of our method with TR-DETR and UVCOM for moment retrieval (MR) and highlight detection (HD) on the QVHighlights \textit{val} set.}
\label{fig3_supp}
\end{figure*}

\newpage

\section*{References}
\small Lei, J.; Berg, T. L.; and Bansal, M. 2021. Detecting moments and highlights in videos via natural language queries. Advances in Neural Information Processing Systems, 34: 11846–11858 \\
\small Sun, H.; Zhou, M.; Chen, W.; and Xie, W. 2024. Tr-detr: Task-reciprocal transformer for joint moment retrieval and highlight detection. In Proceedings of the AAAI Conference on Artificial Intelligence, volume 38(5), 4998–5007. \\
\small Xiao, Y.; Luo, Z.; Liu, Y.; Ma, Y.; Bian, H.; Ji, Y.; Yang, Y.; and Li, X. 2024. Bridging the gap: A unified video comprehension framework for moment retrieval and highlight detection. In Proceedings of the IEEE/CVF Conference on Computer Vision and Pattern Recognition, 18709–18719. \\

\end{document}